\documentclass{article}

    \PassOptionsToPackage{numbers}{natbib}

    \usepackage[preprint]{neurips_2024}

\usepackage[utf8]{inputenc} 
\usepackage[T1]{fontenc}    
\usepackage{hyperref}       
\usepackage{url}            
\usepackage{booktabs}       
\usepackage{amsfonts}       
\usepackage{nicefrac}       
\usepackage{microtype}      
\usepackage{xcolor}         
\usepackage{graphicx}
\usepackage{amsmath}
\usepackage{amsthm}
\usepackage{subcaption} 
\usepackage{adjustbox}
\usepackage{makecell}

\title{From Efficiency to Equity: Measuring Fairness in Preference Learning}

\author{%
  Shreeyash Gowaikar \\
  Department of Computer Science\\
  Birla Institute of Technology and Science, Pilani\\
  \texttt{f20201719@goa.bits-pilani.ac.in} \\
  \And
  Hugo Berard \\
  UNESCO Chair in Urban Landscape \\
  University of Montréal \\
  \texttt{hugo.berard@umontreal.ca} \\
  \And
  Rashid Mushkani \\
  UNESCO Chair in Urban Landscape \\
  University of Montréal  \\
  \texttt{rashid.ahmad.mushkani@umontreal.ca} \\
  \And
  Shin Koseki \\
  UNESCO Chair in Urban Landscape \\
  University of Montréal  \\
  \texttt{shin.koseki@umontreal.ca} \\
}

\begin{document}

\theoremstyle{definition}
\newtheorem{definition}{Definition}[]

\theoremstyle{definition}
\newtheorem*{definition*}{Definition}

\maketitle

\begin{abstract}
As AI systems, particularly generative models, increasingly influence decision-making, ensuring that they are able to fairly represent diverse human preferences becomes crucial. This paper introduces a novel framework for evaluating epistemic fairness in preference learning models inspired by economic theories of inequality and Rawlsian justice. We propose metrics adapted from the Gini Coefficient, Atkinson Index, and Kuznets Ratio to quantify fairness in these models.
We validate our approach using two datasets: a custom visual preference dataset (AI-EDI-Space) and the Jester Jokes dataset. Our analysis reveals variations in model performance across users, highlighting potential epistemic injustices. We explore pre-processing and in-processing techniques to mitigate these inequalities, demonstrating a complex relationship between model efficiency and fairness.
This work contributes to AI ethics by providing a framework for evaluating and improving epistemic fairness in preference learning models, offering insights for developing more inclusive AI systems in contexts where diverse human preferences are crucial.

\end{abstract}

\section{Introduction}
\label{intro}
The rapid advancement of generative artificial intelligence (AI) has brought unprecedented capabilities in natural language processing and content generation. However, these developments have also raised significant concerns about the potential for these systems to perpetuate or amplify epistemic injustice \citep{kay2024epistemic}. Epistemic injustice, a concept introduced by Miranda Fricker~\cite{fricker2007epistemic}, refers to wrongs done to individuals in their capacity as knowers. In the context of generative AI, this manifests as the misrepresentation or misunderstanding of views from minorities and marginalized groups, potentially subjecting them to epistemic violence by denying their own subjective experience.

Generative AI systems, such as large language models, risk producing epistemic injustices by embedding biases through their training data and amplifying narratives from the Global North while silencing voices from the Global South. This imposition of a singular framing onto diverse global perspectives fails to recognize the multitude of views and experiences that exist worldwide. The challenge, therefore, is to develop generative AI systems that can fairly represent all existing views and perspectives, acknowledging and respecting the diversity of human experiences.

To address these alignment challenges, researchers have turned to techniques like Reinforcement Learning with Human Feedback (RLHF) \citep{ziegler2019fine}. RLHF typically involves a multi-step process: first, gathering a dataset where human annotators indicate their preferences among a set of AI-generated options; second, training a reward model on this dataset to predict which options were preferred; and finally, using this reward model to fine-tune the generative model to align it with human preference. While this approach can improve alignment between AI outputs and human values, it raises new questions about epistemic justice.

The concept of the "tyranny of the majority," long recognized in political theory \citep{mill1985liberty}, becomes relevant in this context. If the reward model is biased towards certain groups, it may capture the preferences of dominant groups at the expense of marginalized voices. This scenario could lead to a digital manifestation of majority rule, where the opinions and values of the numerical majority consistently overshadow those of minority groups in AI-generated content.

To mitigate this risk, we argue that reward models should be trained on a diverse set of preferences from annotators who are as representative of the global population as possible. However, even with a diverse dataset, the underrepresentation of minorities may still lead to their perspectives being overlooked in the final model. Therefore, it is crucial to develop methods for measuring and addressing this form of epistemic injustice within reward models.

In this paper, we propose novel metrics, inspired by economic literature on inequality and fair allocation, to quantify the extent to which reward models equally capture the preferences of all users. We demonstrate the application of these metrics on two preference learning tasks, revealing that epistemic injustice can persist even in models with high overall accuracy. Our findings underscore the importance of looking beyond aggregate performance metrics to ensure equitable representation of diverse perspectives.

Furthermore, our research highlights a critical gap in the field: the scarcity of datasets containing individual annotations that would allow for more nuanced analysis of preference distribution across different groups. We advocate for the creation and public release of such datasets, which are essential for advancing research on epistemic justice in AI and developing more inclusive generative models.
By addressing these challenges, we aim to contribute to the development of generative AI systems that not only perform well on standard metrics but also uphold principles of epistemic justice, ensuring that the diversity of human knowledge and experience is respected and accurately represented in AI-generated content.

\section{Related Work} 
\label{litreview}
The intersection of fairness, epistemic justice, and AI has garnered significant attention in recent years, particularly in the context of classification and regression tasks. However, less attention has been paid to fairness and epistemic considerations in more complex AI systems such as generative models and preference learning algorithms.

\paragraph{Fairness in Classification and Regression} In traditional machine learning tasks, fairness metrics often focus on equalizing outcomes across different groups. \cite{hardtEqualityOpportunitySupervised2016} introduced the concept of Equal Opportunity, which aims to equalize true positive rates between protected and unprotected groups. Other measures include equalized odds, ensuring equal probability of positive outcomes across classes, and equal accuracy, which balances performance across groups \citep{cotterOptimizationNonDifferentiableConstraints2019}. These metrics, while valuable, primarily address fairness in standard classification tasks, where the goal is to ensure that the algorithm's output does not depend on sensitive attributes. However, in the context of Reinforcement Learning from Human Feedback (RLHF) and preference learning, the concept of fairness requires a different approach. In these scenarios, we acknowledge that different groups may have varying preferences, and thus, the output of the reward model should rightfully depend on the user. Our notion of fairness in this context focuses on ensuring that the model's accuracy in capturing these diverse preferences does not vary significantly across different groups and individual users.

\paragraph{Fairness in Preference Learning and Ranking} While works like \cite{narasimhanPairwiseFairnessRanking2019} and \cite{beutelFairnessRecommendationRanking2019} have proposed fairness metrics for ranking tasks, their focus primarily remains on ensuring fair treatment of the items being ranked. \cite{sanyalHowUnfairPrivate2022} take a step further by discussing the equity of subgroup populations and exploring the trade-off between this equity and overall accuracy. However, there remains a crucial distinction between these approaches and ours. Standard fairness approaches in ranking typically aim to ensure that items from different groups (e.g., protected vs. unprotected) have equal opportunities to be ranked highly. This focus on the fairness of outcomes for ranked items is important but does not address the full spectrum of fairness concerns in preference learning scenarios.

Our approach, in contrast, shifts the focus to the equity of the participants providing the rankings or expressing preferences. We argue that in preference learning and RLHF contexts, it's crucial to ensure that the model's ability to capture and represent preferences is equitable across all participants, regardless of their group membership. This means that while the content of preferences may vary across groups (which is expected and acceptable), the accuracy with which these preferences are captured and represented by the model should be consistent across all participants.

This distinction is particularly important in scenarios where diverse viewpoints and experiences are critical, such as in collaborative decision-making or in developing AI systems that need to be responsive to a wide range of user preferences. By focusing on the equity of participants rather than just the ranked items, we aim to develop models that are truly representative of diverse human preferences and experiences.

This perspective reveals a critical gap in the literature: most current approaches treat participants as interchangeable, assuming a universal preference or aggregating scores without considering individual differences \citep{sawyerUtilitiesIssueFairness1976,bakker2022finetuning}. This "One-Truth" fallacy \citep{aroyo_truth_2015} fails to account for the diversity of human preferences, which can stem from personal, environmental, and socio-demographic factors \citep{sandri-etal-2023-dont}.

\paragraph{Towards Diverse Preference Representation} Recent work has begun to acknowledge the importance of diverse human preferences in AI alignment and preference learning. Approaches include developing multiple reward models \citep{chakraborty2024maxminrlhf,bakker2022finetuning}, multi-policy strategies \citep{ramé2023rewardedsoupsparetooptimalalignment}, and consensus-based ranking \citep{kovač2023largelanguagemodelssuperpositions}. However, there remains a lack of consistency in how diversity and model performance are measured and evaluated.
Most studies rely on basic classification metrics like F1 score and accuracy across users \citep{sandri-etal-2023-dont,yu2023constructivelargelanguagemodels, zeng2024diversifiedpreferenceslargelanguage} or average reward values \citep{ramé2023rewardedsoupsparetooptimalalignment,bakker2022finetuning}. While these metrics provide some insight, they fail to capture the nuanced ways in which AI systems might perpetuate epistemic injustice by misrepresenting or undermining the subjective experiences of marginalized groups.

\section{Background}
This section introduces a mathematical framework for preference learning that allows us to quantify both the overall performance of a model and its fairness across diverse users.

\paragraph{Problem Definition} Consider a set of $k$ users $\mathcal{U} = \{1, ..., k\}$ and a dataset $\mathcal{D}=\{(x_i, x'_i, s_i, u_i)\}_{i=1}^n$ composed of $n$ pairwise comparisons. Each entry in the dataset represents a comparison where user $u_i$ provided a score $s_i\in \mathcal{S}$. The score can be either: a binary variable (i.e. $\mathcal{S} = \{0, 1\}$ indicating which option was preferred, or a real value (i.e. $\mathcal{S} = \mathbb{R}$), where negative scores indicate preference for $x_i$, and positive scores preference for $x'_i$, and the magnitude of $s_i$ reflects the strength of the preference. Our goal is to learn a model $f: \mathcal{X}\times\mathcal{X} \rightarrow \mathcal{Y}$ that can score any pair $(x,x') \in \mathcal{X}\times\mathcal{X}$. We define the error of the model as: 
$$\mathcal{E}(f) = \mathbb{E}[\ell(f(x_i, x'_i), s_i)]$$
where $\ell$ is a loss function that computes the discrepancy between the model outputs and the ground truth score. The choice of loss function depends on the nature of the scores:
\begin{enumerate}
    \item For real-valued scores, we can use the squared error: $$\ell(f(x_i, x'_i), s_i) = (f(x_i, x'_i) - s_i)^2$$
    \item For binary scores, where $f$ predicts the probability that $x_i$ is preferred over $x'i$, we can use the binary cross-entropy (BCE) loss: $$\ell(f(x_i,x'_i), s_i) = s_i\log(f(x_i,x'_i)) + (1-s_i)\log(1-f(x_i,x'_i))$$
    \item Alternatively, we can use the 0-1 loss: $$\ell(f(x_i,x'_i), s_i) = \begin{cases}
    0 & \text{if } s_i = f(x_i,x'_i) \\
    1 & \text{else}
\end{cases}$$
\end{enumerate}

To evaluate the model's performance for individual users, we define the user-specific error for user $u$: $$\mathcal{E}_u(f) =  \mathbb{E}[\ell(f(x_i, x'_i), s_i) | u]$$

Drawing inspiration from game theory literature on fair allocation, we introduce two key concepts that will help us evaluate both the overall performance and the fairness of our preference learning models.
\begin{definition}[Efficiency] This efficiency measures the overall performance of the model across all users.For a model $f$, it is the mean of the errors across all users: 
$$ \bar{\mathcal{E}}(f) = \frac{1}{|\mathcal{U}|}\sum_{u\in \mathcal{U}} \mathcal{E}_u(f)$$
\end{definition} 

\begin{definition}[Equality] The equality measure helps us quantify the fairness of the model by looking at the worst-case performance. The equality of a model $f$ is the maximum error among all users:  
$$ \mathcal{E}_{\max}(f) = \max_u \mathcal{E}_u(f)$$
\end{definition}

These definitions formalize the trade-off between overall performance and fairness in preference learning models. While traditional machine learning approaches focus on minimizing average error, this can lead to performance disparities across users, potentially perpetuating epistemic injustices. By introducing equality alongside efficiency, we propose a framework for developing models that maintain consistent accuracy across all users.

Our focus on equality, particularly through the $\mathcal{E}_{\max}(f)$ metric, resonates with John Rawls' maximin principle of justice \citep{rawls2017theory, kenfack2024a}. This approach prioritizes the welfare of the worst-off members, which in our context translates to minimizing the maximum error across all users. This Rawlsian perspective provides a philosophical justification for prioritizing outcomes for disadvantaged users or groups, ensuring that AI systems accurately represent all preferences, including those from marginalized or underrepresented populations.

\section{Equality Metrics}
\label{sec:metrics}
We now introduce a comprehensive set of metrics to quantify the extent to which a model's performance varies across users. These metrics, adapted from the economics literature on income inequality, allow us to measure different aspects of fairness and equality in AI systems.

The importance of these metrics lies in their ability to capture various manifestations of inequality in model performance. For instance, errors might be highly dissimilar across specific groups of users or may vary more gradually across the user population. By employing a range of metrics, each with distinct characteristics, we can gain a nuanced understanding of how fair our preference learning models is.
All of the following metrics are non-negative and equal to zero only when the model's performance is identical for all users, representing perfect equality.

\paragraph{Maximal Error Gap}
The Maximal Error Gap measures the largest discrepancy in model performance between any two users. This metric is particularly useful for identifying extreme cases of inequality and aligns with Rawlsian principles of justice by highlighting the worst-case scenario.
    \begin{equation}
        G_{\max}(f) = \max_{u,u' \in \mathcal{U}}(\mathcal{E}_u(f) - \mathcal{E}_{u'}(f))
    \end{equation}

\paragraph{Standard Deviation of the Error} This metric provides a measure of the overall spread of errors across users. A large standard deviation indicates significant variability in model performance, suggesting unequal representation of user preferences.
\begin{equation}
        \sigma^2(f) = \frac{1}{|\mathcal{U}|}\sum_{u \in \mathcal{U}}{(\mathcal{E}_u(f) - \bar{\mathcal{E}}(f))}
    \end{equation}
    
\paragraph{Gini Coefficient \citep{gini1936measure}} The Gini Coefficient, widely used in economics to measure income inequality, it provides a holistic view of error distribution across users. It can be visualized using the Lorenz curve, where the coefficient represents the area between the line of perfect equality and the actual error distribution curve. The Gini Coefficient is bounded between 0 (perfect equality) and 1 (extreme inequality).
\begin{equation}
        G(f) = \frac{\sum_{u, u'}{(|\mathcal{E}_u(f) - \mathcal{E}_{u'}(f)|)}}{2 |\mathcal{U}|^2 \bar{\mathcal{E}}(f)}
    \end{equation}

\paragraph{Generalised Entropy Index \citep{shorrocks1980class}} This index offers flexibility through its $\alpha$ parameter, allowing us to focus on different parts of the accuracy distribution across users. Lower $\alpha$ values are sensitive to the existence of users with low accuracy. While higher $\alpha$ values are more sensitive to the existence of users with high accuracy.
    \begin{equation}
G_\alpha(f)= \begin{cases}

\frac{1}{|\mathcal{U}|} \sum_{u\in\mathcal{U}} \frac{\mathcal{E}_u(f)}{\bar{\mathcal{E}}(f)} \ln \frac{\mathcal{E}_u(f)}{\bar{\mathcal{E}}(f)} & \text{if }\alpha=1 \\
-\frac{1}{|\mathcal{U}|} \sum_{u\in\mathcal{U}} \ln \frac{\mathcal{E}_u(f)}{\bar{\mathcal{E}}(f)} & \text{if }\alpha=0 \\
\frac{1}{|\mathcal{U}| \alpha(\alpha-1)} \sum_{u \in \mathcal{U}}\left[\left(\frac{\mathcal{E}_u(f)}{\bar{\mathcal{E}}(f)}\right)^\alpha-1\right] & \text{else}
\end{cases}
\end{equation}  

\paragraph{Atkinson Index \citep{atkinson1970measurement}} Similar to the Generalized Entropy Index, the Atkinson Index uses an $\epsilon$ parameter to focus on inequalities at different ends of the accuracy distribution across users. As $\epsilon$ increases, the index becomes more sensitive to errors at the lower end of the distribution.
\begin{equation}
A_{\epsilon}\left(f\right)= \begin{cases}

1-\frac{1}{\bar{\mathcal{E}}(f)}\left(\prod_{u \in \mathcal{U}} \mathcal{E}_u(f)\right)^{1 / N} & \text {if } \epsilon=1 \\ 
1-\frac{1}{\bar{\mathcal{E}}(f)} \min_{u \in \mathcal{U}} \mathcal{E}_u(f)  & \text {if } \epsilon=+\infty \\
1-\frac{1}{\bar{\mathcal{E}}(f)}\left( \frac{\sum_{u \in \mathcal{U}} \mathcal{E}_u(f)^{1-\epsilon}}{|\mathcal{U}|}\right)^\frac{1}{1-\epsilon} & \text{else}
\end{cases}
\end{equation}

\paragraph{Kutznets Ratio \citep{kuznets2019economic}}
Unlike the previous metrics that consider the entire error distribution, the Kuznets Ratio focuses on the extremes, comparing the errors of the top $\alpha\%$ of users to the bottom $\alpha\%$. This metric is particularly useful for identifying disparities between the best and worst-served users:
\begin{equation}
    \label{kutznets}
        K_{\alpha}(f) = \frac{\sum_{\text{top } \alpha\%} \mathcal{E}_u(f)}{\sum_{\text{bottom } \alpha\%} \mathcal{E}_u(f)}
    \end{equation}

By employing this diverse set of metrics, we can comprehensively evaluate the fairness and equality of preference learning models, allowing us to focus on different ends of the distribution.

\section{Experimental Setup}
\label{methods}

To rigorously evaluate our proposed equality metrics, we carefully selected two datasets that provide crucial individual-level annotation data. This granular information—detailing which user provided each annotation—enables us to precisely analyze variations in model performance across diverse users. Such user-specific data is instrumental in uncovering potential epistemic injustices in AI systems, yet it is often absent from many widely-used datasets in the field of Reinforcement Learning from Human Feedback (RLHF).

Notably, prominent datasets such as Safe RLHF \citep{dai2024safe}, Helpful and Harmless \citep{bai2022traininghelpfulharmlessassistant}, WebGPT \citep{nakano2022webgptbrowserassistedquestionansweringhuman}, ImageReward \citep{xu2023imagereward}, and AVA \citep{avadata} lack user identification data. This omission precludes the differentiation of users with diverse preferences, rendering the computation of our proposed equality metrics impossible on these datasets. The absence of such critical information in these widely-used resources highlights a significant gap in the field's ability to assess and address epistemic injustice in generative models.

We posit that the inclusion and release of user-specific annotation data is not merely beneficial but essential for the comprehensive evaluation of fairness in AI systems. By enabling the application of metrics like those proposed in this study, such data would significantly enhance our capacity to identify, quantify, and ultimately mitigate epistemic injustices in generative models. This underscores the urgent need for more nuanced and comprehensive datasets in the pursuit of truly fair and equitable AI systems.

To ensure robustness of the results, all experiments employ the following methodology: 1) Model selection was based on the best performance on a randomized validation set, using comparisons not present in the training data. 2) Each model was independently trained five times to account for stochasticity and we report the average results across the five runs and the corresponding confidence intervals.

\subsection{AI-EDI-Space Dataset}
\begin{figure}[h]
\centering
\includegraphics{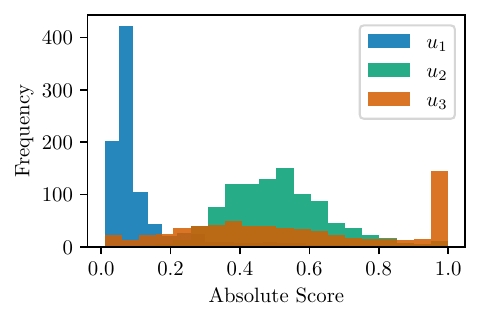}
\caption{Voting patterns observed in the AI-EDI-Space Dataset. We plot the histogram of the absolute scores given by 3 participants with clearly distinct voting patterns. This figure highlights that users might have very different "taste profiles".}
\label{voting_patterns}
\end{figure}

\paragraph{Dataset} The first dataset consists of 7,833 street-view images representing a diverse set of public spaces from the Montreal Metropolitan Area. The dataset includes 19,990 pairwise comparisons, evaluated by 22 individuals who were carefully selected to maximize diversity and include underrepresented groups based on ethnicity, gender, sexuality and age. Each participant evaluated a minimum of 500 comparisons based on 35 different criteria designed to capture various qualities of public spaces. Participants provided a real value between -1 and 1 to avoid Arrow’s impossibility theorem \cite{arrowDifficultyConceptSocial1950}, as explained by \cite{allouahRobustSparseVoting2022}. However, this approach introduces complexity into the voting patterns, as illustrated in Figure \ref{voting_patterns}.

This dataset is ideal for testing algorithmic equity, as it includes a diverse set of participants with potentially divergent views on what makes a public space valuable, making the task highly subjective.

\paragraph{Model} The model and training procedure used are similar to the one proposed in \cite{dubeyDeepLearningCity2016}. The model takes a single image as input and predicts the scores. The model consists of a feature extractor and a classifier head.
The feature extractor is a pre-trained model. We experimented with several models, including VGG11 \cite{simonyanVeryDeepConvolutional2014}, EfficientNet \cite{tanEfficientNetRethinkingModel2020}, Squeezenet \cite{iandolaSqueezeNetAlexNetlevelAccuracy2016}, and DinoV2 \cite{oquabDINOv2LearningRobust2023}. 
The extracted features were then passed through a classifier head to predict a score for each of the 35 criteria.
We observed that the model with EfficientNet, along with a double-layered classification head with 256 as the hidden dimension, gave the best results throughout all the experiments. 
Hence, all values for the AI-EDI-Space dataset are using an EfficientNet feature extractor.
To train the model, we computed the scores for both images in each comparison and calculated the difference in scores between the two images. We then used the Mean Squared Error between the difference in scores and the ground truth score as the error to train the model.

\paragraph{Loss} We use the 0-1 loss as the baseline metric to compute the various equality metrics, which corresponds to measuring the difference in accuracy across users.

\subsection{Jester Jokes Dataset}
\paragraph{Dataset} We also test the proposed metrics on the Jester Jokes Dataset, originally developed for recommender systems research \citep{goldbergEigentasteConstantTime2001}. 
This dataset contains 100 jokes, each rated on a scale from -10 to +10 by 73,421 participants. The inherently subjective nature of humor appreciation makes this dataset particularly suitable for examining model performance across diverse user preferences. To ensure data quality and diversity, we only selected a subset of the annotations. We first filtered the dataset to include only participants who rated all 100 jokes, ensuring comprehensive engagement from each user. From this filtered set, we then selected 1,000 users exhibiting the most diverse voting patterns. This selection was achieved through Principal Component Analysis (PCA) of users' score vectors, followed by a uniform sampling of users maximally distant in the PCA space. This careful curation process resulted in a dataset that not only maintains balance during model training but also captures a wide spectrum of user preferences.

\paragraph{Model} Since users directly provided scores for each joke, the problem can be formulated as a regression task. We trained a model that takes a single joke as input and predicts its score. The model consists of a feature extractor and a classifier head. The feature extractor is a pre-trained BART model \cite{lewis-etal-2020-bart}. The extracted features are then passed through a classifier head, which is either a single-layer or a double-layer perceptron, to predict the score. The Mean Squared Error (MSE) between the predicted score and the ground truth score  was used as the loss function to train the model.

\paragraph{Loss} We used the MSE loss as the baseline metric to compute the various equality metrics.

\section{Methods}
\label{solutions}
To address the challenge of inequality in model performance across users, we propose and evaluate two categories of techniques: pre-processing and in-processing. These approaches aim to enhance the fairness of preference learning models by ensuring more equitable representation of diverse user preferences.

\subsection{Pre-Processing Techniques}
Pre-processing techniques are applied to the data prior to model training. We investigate three scaling methods designed to normalize the distribution of scores across users:
\begin{enumerate}
    \item \textbf{Min-Max Scaling}: This technique scales each participant's scores to a range of [-1, 1]. It is applied individually to each user's scores, preserving relative preferences within a user's data while enabling comparability across users.
    \item \textbf{Normalization Scaling}: This two-step process first applies standard normalization to each participant's scores, adjusting the mean to 0 and standard deviation to 1. Subsequently, the scores are scaled to ensure they remain within the [-1, 1] range. While this method, like Min-Max Scaling, is user-specific, it does not guarantee sparse unanimity.
    \item \textbf{Mehestan Scaling \cite{allouahRobustSparseVoting2022}}: This more sophisticated approach considers the voting patterns of all participants when scaling an individual's scores. The process involves: a) Converting raw comparison scores to individual scores using a Generalized Bradley-Terry Model \citep{fageotGeneralizedBradleyTerryModels2023}. b) Scaling and translating these scores using the BrMean primitive, which is designed to be resilient to potential manipulation by malicious voters. c) Preserving individual score distributions without final aggregation, maintaining the uniqueness of each participant's preference pattern.
\end{enumerate}

Mehestan Scaling is particularly effective in achieving sparse unanimity, a property that ensures the preservation of unanimous preferences even when user voting patterns differ significantly.

\subsection{In-Processing Techniques}
In-processing techniques are integrated into the model training process. We explore two primary approaches:
\begin{enumerate}
    \item \textbf{User Embeddings}: By incorporating user-specific embeddings as additional input to the model, we aim to capture and adapt to individual voting patterns. This technique allows the model to learn user-specific features that may influence preference judgments.
    \item \textbf{Contrastive Loss}: We employ contrastive loss in conjunction with least squares error (LSE). The contrastive loss works by increasing the distance between dissimilar scores, ensuring that the model's outputs are not clustered too closely together. This helps prevent comparisons—calculated as the difference between the scores of two alternatives—from being too close to zero, thereby improving the model's ability to distinguish between different preferences.
\end{enumerate}

\section{Results}
\label{results}

\begin{figure}[h]{
    \adjustbox{margin=-1.5cm 0cm -0.5cm 0cm}{ 
    \subcaptionbox{Sorted Per User Accuracy over i) all Criteria, ii) Top 3 Criteria - Opperessing, Vegetated/Green, Refreshing (in the descending order of accuracy). \label{reg_res}}{%
        \includegraphics[width=0.35\textwidth]{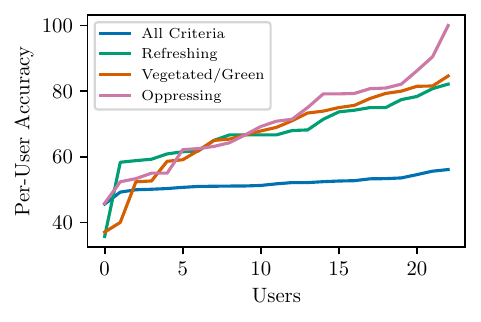}} 
    \hspace{0.5cm}
    \subcaptionbox{Sorted Per User Accuracy over i) All Criteria, ii) only 5 Criteria with highest overall accuracies (viz. Intimate, Regenerative, Refreshing, Vegetated/Green, Oppressing).  \label{top_5_criteria_res}}{%
        \includegraphics[width=0.35\textwidth]{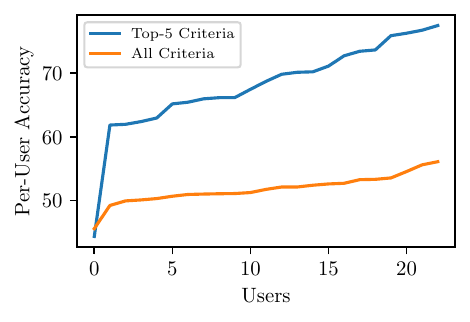}} 
    \hspace{0.5cm} 
    \subcaptionbox{Scatter Plot of accuracies for every criterion. The highest accuracy is for the criterion 'Opperessing', and the lowest accuracy is for the criterion 'Inviting/Welcoming'. \label{criteria_res}}{%
        \includegraphics[width=0.35\textwidth]{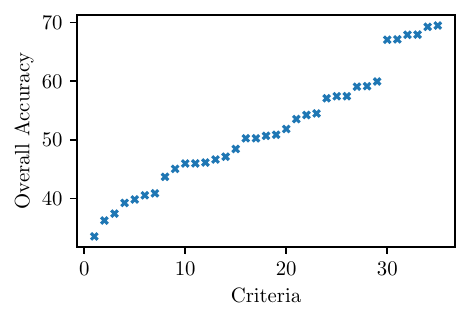}} 
    }
    }
    \caption{Experiment results on the AI-EDI Space Dataset}
    \label{fig_results}
\end{figure}

\paragraph{AI-EDI-Space Dataset} After training a model on the AI-EDI-Space dataset, we computed the various equality metrics proposed earlier. Figure~\ref{fig_results} shows the distribution of accuracy for all users across different criteria. Our analysis yielded several noteworthy observations: inequality appears to be higher for criteria where the model performs best overall, as shown in Table~\ref{sol_results}. This illustrates the potential trade-off between efficiency and equality.
The observed inequality seems to be primarily driven by voting patterns, as the mean squared error (MSE) used to train the model penalizes discrepancies with the user's comparisons. 
We also observe that users whose voting patterns cluster around 0 (a conservative voting approach) tend to achieve higher accuracy.
Additionally, the number of comparisons annotated by each user may differ, potentially contributing to the observed inequalities. 
However, disentangling these effects requires further analysis to understand the influence of sample size on our observations.

\begin{table*}[]
    \adjustbox{margin= -1.2cm 0cm -0cm 0cm}{
    \begin{tabular}{lccccccc}
        \toprule
         Experiment & Accuracy & $G_{max}$  & $\sigma^2$ & \thead{$G$\\$\times10^{-2}$} & \thead{$G_{\alpha=0}$\\$ \times 10^{-4}$} & \thead{$A_{\epsilon=\infty}$\\$\times10^{-1}$} & $K_{\alpha = 20}$
        \\
        \midrule
         Normalisation & $51.1 \pm 1.9$ & $6.6 \pm 1.1$ & $1.7 \pm 0.2$ & $0.9 \pm 0.1$ & $5.5 \pm 1.5$ & $0.7 \pm 0.1$ & $1.37 \pm 0.02$
         \\
         MinMax & $50.2 \pm 2.6$ & $7.6 \pm 4.0$ & $1.8 \pm 0.8$ & $1.0 \pm 0.4$ & $7.1 \pm 5.8$ & $0.7 \pm 0.6$ & $1.38 \pm 0.03$
         \\
         Mehestan & $51.3 \pm 4.4$ & $7.1 \pm 1.2$ & $1.8 \pm 0.3$ & $1.1 \pm 0.2$ & $7.4 \pm 3.4$ & $0.7 \pm 0.2$ & $1.39 \pm 0.04$ 
         \\
         Contrastive Loss & $52.0 \pm 0.6$ & $7.7 \pm 1.8$ & $2.1 \pm 0.6$ & $1.1 \pm 0.3$ & $8.0 \pm 4.7$ & $0.8 \pm 0.1$ & $1.40 \pm 0.04$ 
         \\
         User Emb. & $50.2 \pm 0.7$ & $7.8 \pm 1.0$ & $1.9 \pm 0.4$ & $1.0 \pm 0.2$ & $7.0 \pm 2.6$ & $0.8 \pm 0.4$ & $1.39 \pm 0.03$ 
         \\
    \bottomrule
                 
    \end{tabular}}
    \centering
    \caption{Results for the AI-EDI Image Dataset over different experiments. Individual Users evaluated using Per-User Accuracy}
    \label{sol_results}
\end{table*}

\begin{figure}[h] 
    \centering 
    \includegraphics[width=0.5\textwidth]{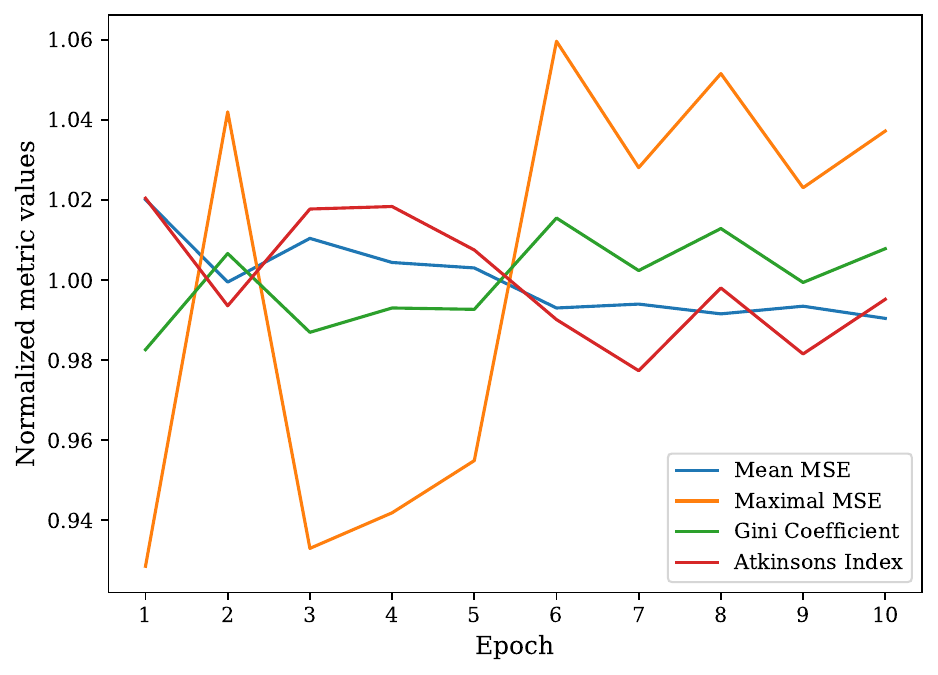} 
    \caption{Performance and Metrics for Jester Jokes Dataset over epochs. A clear tradeoff in performance and equality can be seen from the figure where equality seems to increase over training epochs and performance decreases.} 
    \label{fig:tradeoff} 
\end{figure}

\paragraph{Jester Jokes Dataset} Our analysis of the Jester Jokes dataset provided additional insights into the effectiveness of various scaling and translation techniques in addressing inequality. Table~\ref{sol_results_jester} presents the results of our equality metrics for different approaches. We made the following observatoins: 1) Normalization and MinMax Scaling achieved low Mean Squared Error (MSE) but failed to significantly improve equality. Notably, the Kuznets ratio remained close to 4, indicating substantial inequality between the top 20\% and bottom 20\% of participants in terms of model performance. 2) Mehestan Scaling, designed for sparse unanimity \cite{allouahRobustSparseVoting2022}, yielded higher equality values despite a worse MSE. This finding suggests that techniques prioritizing unanimous preference recovery may contribute to greater epistemic fairness. 3) Given our careful selection of users who scored all 100 jokes, we can primarily attribute the observed inequity to differences in voting patterns rather than data imbalance. This reinforces the importance of considering diverse preference expressions in model development.
Additionally, we can clearly see the tradeoff between performance and equality from Figure~\ref{fig:tradeoff} where the average MSE falls over training epochs, indicating performance gains, whereas the Gini coefficient and Maximal MSE increase, indicating increased inequality. 
 
\begin{table*}[]
    \adjustbox{margin=-1.7cm 0cm -1.7cm 0cm}{
    \begin{tabular}{lccccccc}
        \toprule
         Experiment & \thead{$MSE$ \\$\times 10^{-1}$} & \thead{$G_{max}$\\$\times 10^{-1}$}  & \thead{$\sigma^2$\\$\times 10^{-1}$} & \thead{$G$\\$\times 10^{-1}$} & \thead{$G_{\alpha=0}$\\ $\times 10^{-1}$} & \thead{$A_{\epsilon=\infty}$\\$\times 10^{-1}$} & $K_{\alpha = 20}$
        \\
        \midrule
         Normalisation & $2.66 \pm 0.03$ & $7.6 \pm 0.3$ & $1.38 \pm 0.03$ & $1.43 \pm 0.02$ & $1.55 \pm 0.04$ & $9.6 \pm 0.1$ & $4.81 \pm 0.08$
         \\
         MinMax & $2.61 \pm 0.01$ & $7.8 \pm 0.5$ & $1.35 \pm 0.01$ & $1.43 \pm 0.01$ & $1.55 \pm 0.02$ & $9.6 \pm 0.3$ & $4.84 \pm 0.04$
         \\
         Mehestan & $8.86 \pm 2.12$ & $21.9 \pm 5.0$ & $3.37 \pm 0.98$ & $1.02 \pm 0.01$ & $0.67 \pm 0.15$ & $6.7 \pm 0.4$ & $2.81 \pm 0.37$ 
         \\
        
    \bottomrule
    
    \end{tabular}}
    \caption{Results for the Jester Jokes Dataset over different experiments. Individual Users evaluated using Per-User Mean Squared Error}
    \label{sol_results_jester}
\end{table*}

\section{Conclusion}
\label{concl}
In the era of generative AI, where algorithms increasingly shape decision-making processes, ensuring that these systems do not generate or amplify epistemic injustice is paramount. This study introduces a novel perspective on fairness in preference learning, focusing on the equitable representation of diverse human preferences and views. Our work provides a framework for quantifying and addressing epistemic fairness in AI models, contributing to the development of more just and inclusive AI technologies.
Our findings underscore the potential for significant disparities in how well AI models capture preferences across different users. This raises critical questions about epistemic justice in AI systems and highlights the need for further research in several key areas: 1) Further investigation is needed to understand the sources of inequality in model performance and develop effective mitigation strategies. This includes examining how data characteristics, model architectures, and diverse human preferences interact to produce or exacerbate inequalities. 2) As Reinforcement Learning from Human Feedback (RLHF) becomes more prevalent, ensuring that the alignment process itself is equitable across diverse participant groups is crucial. This involves developing methods to capture a wide range of opinions and preferences, particularly from marginalized or underrepresented groups. 3) We advocate for the public release of more datasets that include annotation-level user information, specifically detailing which annotator or user provided each individual annotation. This granular data is crucial for conducting comprehensive evaluations of epistemic justice in AI models. Such datasets would enable researchers to track how different users' preferences and judgments are represented in model outputs, providing a more nuanced understanding of potential biases or inequalities in preference learning and generative AI systems. 4) Finally, the tension between individual fairness and overall system efficiency raises important ethical questions. Future work should explore how to balance these concerns in line with principles of distributive justice and epistemic fairness, particularly in the context of generative AI systems.

This study represents a step towards more equitable AI systems that respect and accurately represent diverse human preferences. As AI continues to play an increasingly significant role in society, addressing epistemic fairness will be crucial in ensuring that these systems serve all members of society equitably. Our work provides a foundation for future research in this critical area, aiming to develop AI technologies that are not only efficient but also just and inclusive.

\bibliographystyle{plain}
\bibliography{neurips_2024}

\end{document}